# Robust Anomaly Detection for Time-series Data

**Min Hu [1,2], Yi Wang [1,2], Xiaowei Feng [1,2], Shengchen Zhou [1,2], Zhaoyu Wu [3], Yuan Qin [3]**

[1]SHU-UTS SILC Business School, Shanghai University, Shanghai, China
[2]SHU-SUCG Research Centre for Building Industrialization, Shanghai, China
[3]Shanghai Tunnel Engineering Co., Ltd, Shanghai, China

Corresponding author: Yi Wang (e-mail: wiyixi@163.com).

**ABSTRACT** Time-series anomaly detection plays a vital role in monitoring complex operation conditions. However, the detection accuracy of existing approaches is heavily influenced by pattern distribution, existence of multiple normal patterns, dynamical features representation, and parameter settings. For the purpose of improving the robustness and guaranteeing the accuracy, this research combined the strengths of negative selection, unthresholded recurrence plots, and an extreme learning machine autoencoder and then proposed robust anomaly detection for time-series data (RADTD), which can automatically learn dynamical features in time series and recognize anomalies with low label dependency and high robustness. Yahoo benchmark datasets and three tunneling engineering simulation experiments were used to evaluate the performance of RADTD. The experiments showed that in benchmark datasets RADTD possessed higher accuracy and robustness than recurrence qualification analysis and extreme learning machine autoencoder, respectively, and that RADTD accurately detected the occurrence of tunneling settlement accidents, indicating its remarkable performance in accuracy and robustness.

## I. INTRODUCTION

Increasing data volume and structural complexity impact and pose challenges to the traditional data analysis process but also provide opportunities to deepen the cognition of data [1]. Complex operation conditions and other hidden information in data have been gradually discovered and applied in the fields of condition monitoring and anomaly detection. Detecting abnormal data, especially abnormal time-series data, has become an urgent demand in various domains, such as finance, engineering, and medical treatment [2-5].

An anomaly is generally considered to be a pattern that differs from normal expected behavior [6, 7]. As normal expected behavior is not specific to application situations, there is no uniform standard for anomalies [8]. Therefore, the detection of anomalies relies heavily on the information provided by given data, most of which are normal. Depending on whether the given data contain data labels, anomaly detection methods are broadly classified as supervised or unsupervised learning methods. Supervised learning methods extract features from labeled data to identify the data with abnormal features. For example, Yin *et al.* extracted the spatio-temporal features of time series by a convolutional neural network (CNN), and classified them into normal or abnormal time series with a convolutional autoencoder [9]. However, anomalies count for a small proportion in real data, and there are numerous causes of anomalies, which makes it difficult to obtain typical anomalous features, especially those of time-series data [10]. Therefore, in practical applications, most time-series anomaly detection methods adopt unsupervised learning methods that do not require data labeling. Unsupervised learning methods are inspired by anomaly definition, which involves identifying abnormal patterns that do not conform with the assumed normal time-series data. For example, Mehrang *et al.* detected abnormal time-series data that were not consistent with normal expectations by Autoregressive Integrated Moving Average model, Multivariate Alteration Detection, and Rosner statistical metrics [11]. The main difference between the two time-series anomaly detection types is about normality assumption. The supervised learning method directly learns the features and patterns from labeled data, while the unsupervised anomaly detection method based on unlabeled time-series data needs to first assume that the data before the detection time point are normal, and then judge whether the data after it are normal.

The core of the anomaly detection task is similarity measurement. There are two methods for time-series anomaly detection with respect to similarity measurement: feature-based classification methods and pattern-based prediction methods. Feature-based classification methods directly divide the normal and abnormal data according to the features extracted by statistical indicators or machine learning methods. For instance, Talagala *et al.* extracted 14 statistical features and recognized abnormal time subsequence with obvious changes in feature density [12]. Considering that data distribution might change over time, Dani *et al.* set a dynamic anomaly threshold according to the local mean and standard deviation of the time window [13]. Wang *et al.* used the support vector machine and density-based feature enhancement technology to detect specific types of network intrusion anomalies [14]. Ding *et al.* proposed a similarity measurement method based on a CNN to mine the characteristics of data and mark them in the process of clustering to assist in clustering and improve the effect of anomaly detection [15]. The methods based on pattern-based prediction tend to be unsupervised learning. These methods learn the dynamic pattern and output the prediction results in line with the pattern. For example, Munir *et al.* directly predicted the next time data by a CNN and then detected anomalies with the predicted residual [16]. To strengthen the learning ability for time-series data, Li *et al.* extracted features based on extended fuzzy C-means clustering and detected anomalies of multi-dimensional time-series data through reconstruction error [17]. Wu *et al.* increased the dimension of time-series characteristics by time decomposition and used the stacked Gate Recurrent Unit network to detect anomalies in online time-series data [18].

However, there are four challenges within anomaly detection in current research and applications: unbalanced pattern distribution, existence of multiple normal patterns, difficulty in representing dynamical features, and difficulty in parameter setting. First, as we mentioned before, notwithstanding their minority, abnormal pattern data are various, which cannot be considered completely in multi-classification methods. This imbalance distribution affects the accuracy of unsupervised anomaly detection. Second, traditional time-series anomaly detection methods only consider one normal pattern. However, with the passage of time, time series may contain more than one normal pattern. Pattern misjudgment will occur once the pattern is switched, resulting in high false positives. In addition, both feature-based classification methods and pattern-based prediction methods rely on outstanding feature representation. However, time-series data are nonlinear, and their characteristics are difficult to measure effectively[19]. Therefore, how to represent the hidden dynamical features of time-series data is a difficult problem in time-series anomaly detection. Finally, anomaly detection methods face a common problem: multiple-parameter setting. An improper parameter setting strategy affects the robustness and efficiency of the algorithm to a great extent, which becomes more serious in pattern-based prediction methods [10].

According to the abnormal pattern recognition assumption, this study applied negative selection to effectively use given data. Negative selection can generate detectors based on a self-set where a small amount of labeled data exists, and can recognize abnormal data inconsistent with the self-set [20]. However, the traditional negative selection detector may misjudge unknown normal patterns. Multiple labeled normal patterns in the self-set are incomplete in real-world applications, and there may be other normal patterns over time [21]. Therefore, this study made some improvements based on negative selection. Focusing on the dynamical features representation, this study selected the unthresholded recurrence plot to represent time-series data in the phase space so as to mine the internal dynamical features within time-series data rather than surface features. The unthresholded recurrence plot (URP) is a technique for analyzing complex dynamic systems [22]. It can generate phase-space trajectories of time series and remove the threshold of binary transformation when representing the similarity of phase-space trajectory point pairs. The transformation of the URP not only shows the characteristic mode of the dynamic system in time series but also avoids the loss of potentially useful information. To identify abnormal features in unthresholded recurrence plots, this study used the reconstruction error of the autoencoder to distinguish normal and abnormal mode patterns. The autoencoder takes the input information as the learning goal and can automatically learn the representation of the input information without any data label. At the same time, to effectively learn URPs and reduce the influence of parameter setting, this study further introduced a variant of extreme learning machine (ELM) called the extreme learning machine autoencoder (ELM-AE) [23]. The ELM-AE determines the weights of hidden layer nodes randomly and calculates the output weights by taking the input information as the learning target. It not only possesses the self-learning ability of the autoencoder, which performs well in image feature extraction, but can also conduct fast pattern learning with high generalization performance [24].

Therefore, based on negative selection, URP, and ELM-AE, this study proposed robust anomaly detection for time-series data (RADTD). This method alleviates label dependency, highlights time-series dynamical features, and quickly adapts to high-frequency changes of time-series data. In general, it reduces parameter numbers and mitigates the influence of parameter setting, which requires manual intervention. Besides, by storing and updating the self-set, RADTD can automatically learn the time-series patterns and reduce false alarms during the monitoring process.

The remainder of this article is organized as follows: Section II presents related works about the negative selection, URP method, and ELM-AE for anomaly detection. The proposed framework is introduced in Section III, illuminating the main idea and implementations. The next two sections describe the framework through benchmark and real-world datasets. In Section IV, we describe the comparisons of RADTD with recurrence qualification analysis (RQA) and ELM-AE in accuracy and robustness respectively. In Section

V, we describe the three engineering datasets that were simulated to evaluate the RADTD's accuracy and robustness in the real world. The last section concludes with general remarks and recommendations for further research.

## II. RELATED WORK

In this study, the relevant theoretical knowledge is briefly introduced: negative selection, URP, and ELM-AE.

### A. NEGATIVE SELECTION

Negative selection is a maturation mechanism of T cells in the biological immune system. Its main purpose is to cull T cells that strongly bind self-peptides, ensuring that T cells cannot recognize themselves. Fig. 1 shows a schematic diagram of the negative selection of T cells in the biological immune system. In 1994, negative selection was introduced into artificial immune theory for self and non-self pattern recognition [20]. The main idea of negative selection consists in recognizing non-self patterns and rejecting self patterns by defining the normal dataset as the self-set. Unlike other artificial intelligence algorithms that mainly rely on positive and negative labels, the negative selection stores the characteristic information of self patterns, which contributes to identifying the non-self pattern and alleviates the dependence on negative samples [25]. The negative selection algorithm has greatly promoted the research and application of artificial immune system anomaly detection, such as network intrusion, structural health monitoring, stock market anomaly monitoring, and traffic flow monitoring [26-29]. Therefore, this study used negative selection in pattern recognition to store pattern information so as to identify abnormal patterns.

### B. UNTHRESHOLDED RECURRENCE PLOT

A recurrence plot (RP) is a graphical representation for time series. It was proposed by Eckmann *et al.* to analyze the nonlinear characteristics of time series from the perspective of a dynamic system [30]. The RP technique reconstructs the time series into a high-dimensional phase space and then represents it as a two-dimensional square matrix, where columns and rows axes represent time. In this process, the method marks roughly the same region in the phase-space trajectory of the dynamic system, that is, the recurrence behavior within the time series. Therefore, by analyzing the pattern and texture characteristics in a two-dimensional square matrix, researchers can understand the evolution and recursion of time series and can analyze the periodicity, chaos, and non-stationarity of the time series by a dynamic system.

The transformation from time series into phase-space trajectories by an RP can extract the implicated dynamic features. For example, Waxenegger *et al.* captured the unstable oscillation of rocket combustion by RP analysis [31]. Ozkok and Celik extracted RP features from high-resolution melting curve data and successfully identified different yeast species [32]. Alghawali detected a variety of intrusion types in traffic data based on RP [33]. However, effective RP analysis depends on a proper RP threshold relevant to similarity measurement, and an inappropriate threshold will result in information loss. However, how to optimize the threshold is still under debate. As a variant of an RP, an URP cancels the RP threshold so as to fully represent dynamical information through a phase-space trajectory without information loss, and obtain details of dynamical features. Fig. 2 shows the representation results of the URP and RP for the same temporal time series. The URP applications have concentrated on image representation and demonstrated the dynamical feature extraction ability in pattern recognition. For example, Oliver and Aldrich converted the time series of grinding circuit variables into URPs, which contributed to accurately identifying the mode changes of a grinding circuit [21]. Kok and Aldrich extracted texture features from the URP of electrochemical noise and generated a classifier for corrosion recognition [34]. Therefore, in this study, a URP was applied for the phase-space transformation of the time series to present the dynamical features of the time series.

### C. EXTREME LEARNING MACHINE AUTOENCODER

The autoencoder (AE) is a representative learning method based on a neural network [35, 36]. By setting the input layer data as the target, the encoder of an AE can automatically learn data pattern features from unlabeled data. The process whereby the decoder reconstructs the target from the potential vector also provides convenience for data-driven abnormal pattern recognition.

The extreme learning machine (ELM), proposed by Huang *et al.*, is a method based on the single hidden layer feedforward neural network [37]. Unlike the general BP neural network, ELM replaces the time-consuming hyper-parameter optimization process with random projection, where the input weight and bias are determined once. Besides, the output weight can be computed directly by solving equations without an iterative adjustment. These properties equip ELM with fast representative learning and high learning accuracy.

The extreme learning machine autoencoder (ELM-AE) is an AE based on the ELM, as shown in Fig. 3 [37]. The ELM-AE combines the characteristics of AE and ELM and has been shown to possesses a strong learning ability to distinguish images, even better than traditional PCA analysis [38]. In a practical scenario, Ahmad *et al.* adopted the improved ELM-AE in classifying effective complex hyperspectral data with less training samples, which further reflects the efficient learning ability of ELM-AE for graphics [39]. Given the automatic representative learning ability of the ELM-AE for graphic features, the ELM-AE was chosen in this study to learn RP pattern recognition.

## III. A ROBUST ANOMALY DETECTION FOR TIME-SERIES DATA

A robust anomaly detection for time-series data (RADTD) was proposed. This section first presents the idea of the framework and then describes specifics on each step in the framework implementation.

### A. THE IDEA OF THE RADTD FRAMEWORK

In RADTD, the processing flow of the time series mainly includes three functions: feature extraction, ELM-AE encoding, and anomaly detection, as shown in Fig. 4. For the feature extraction function, RADTD extracts temporal features by converting time series into URPs. During the conversion, the time series are first transformed into a higher-dimensional phase space, highlighting characteristics of the time series from the perspective of dynamics, and then the phase space is described by the two-dimensional square matrix URP, which is also conducive to the rapid learning of the dynamical time-series characteristics. In the ELM-AE encoding function, RADTD uses the ELM-AE to automatically and quickly learn the features of URPs, which are stored in the form of weight parameters for following URP reconstruction. In the anomaly detection function, ELM outputs the reconstructed URP and compares it with the actual URP. The difference between the two URPs is measured by the reconstruction error, and then RADTD can judge whether there are obvious differences in time-series patterns.

In terms of the implementation process, RADTD's anomaly recognition process (Fig. 5) mainly includes five modules: image conversion, auto-encoding training, self-sets, ELM testing, and the anomaly recognition module. Assuming that the historical time series is normal, RADTD can identify the anomalies of following untested time series subsequence. First, the image conversion module converts the historical time series into multiple groups of URP representation to highlight the dynamical features of time series. Then, the auto-encoding training module automatically learns the features of each URP set. The self-set module stores the learned features as a normal time-series pattern. It is remarkable that the negative selection detectors in RADTD can directly apply in recognition based on self-sets, rather than be generated randomly and discarded as in traditional negative selection methods. Besides, data labels are utilized in self-set improvement. Generally, the unlabeled patterns in self-sets are set to normal by default, and to confirm the normality, RADTD also allows us to add labeled patterns, which develop the recognition efficiency. Before the anomaly recognition module becomes effective, untested time-series data are also converted into a URP through the feature extraction module. Then, the ELM testing module, charging for URP reconstruction, tries to reconstruct the URP by the stored corresponding pattern features in self-sets. Finally, RADTD in the anomaly recognition module calculates the reconstruction error for each pattern in the self-set and sets the pattern with the minimum reconstruction error as the matching pattern for testing the URP. Hence, if the reconstruction error is over a certain range, the testing URP will be considered as an anomaly.

### B. IMPLEMENTATION OF THE RADTD

RADTD consists of the following five modules: image conversion, ELM-AE training, self-sets, ELM testing, and anomaly recognition. Specifics for the five modules are described in this section.

#### 1) IMAGE CONVERSION MODULE

Given the $d$-dimensional time series $D = \{\vec{x}_1, \vec{x}_2, ..., \vec{x}_i, ..., \vec{x}_N\}$, where $\vec{x}_i$ represents the $i$-th observation, the framework separates it into subsequences and then converts it into URPs. To eliminate the effect of dimension and magnitude, each dimension of the dataset is first rescaled by (1):

$$\vec{z}_i = \frac{\vec{x}_i - \vec{x}_{min}}{\vec{x}_{max} - \vec{x}_{min}} \quad (1)$$

where $\vec{x}_{min}$ and $\vec{x}_{max}$ are the minimum and the maximum values for dimensions, respectively, and $\vec{z}_i$ is the rescaled value of the $i$-th observed sample. Thus, the rescaled time series is represented as $D' = \{\vec{z}_1, \vec{z}_2, ..., \vec{z}_i, ..., \vec{z}_N\}$. $D'$ is separated into time-series subsequences of length to represent temporal trends in different periods; that is $D' = \{D_1, D_2, ..., D_i, ..., D_T\}$, where $T$ is equal to $N - w + 1$, and $D_i = \{\vec{z}_i, \vec{z}_{i+1}, ..., \vec{z}_{i+w-1}\}$ is the $i$-th time-series subsequence.

Then, RADTD converts the subsequence $D_i$ into a URP for dynamical feature representation. Using the time-delay embedding technique, $D_i$ is reconstructed into the $(w-1-(m-1)\tau, m)$ phase-space array, and its $i$-th phase-space vector is embedded by (2):

$$\begin{aligned} D_{i,j} &= \left\{\vec{z}_{i+j}, \vec{z}_{i+j+\tau}, ..., \vec{z}_{i+j+(m-1)\tau}\right\}, \\ j &\in (1, w-1-(m-1)\tau) \end{aligned} \quad (2)$$

where $m$ is the embedding dimension, and $\tau$ is the time delay. Whereas embedding dimension $m$ has little effect on URP's dynamical features, experiments in this study set $m = 1$.

The recurrence plot is constructed by forming the recurrence matrix $R^i$ by (3),

$$R^i_{j_1, j_2} = \left\| D_{i, j_1} - D_{i, j_2} \right\|, j_1, j_2 = 1, ..., w-1 \quad (3)$$

where $D_{i, j_1}, D_{i, j_2} \in \{D_{i,j}\}_{j=1}^{w-1-(m-1)\tau}$, and $\|\cdot\|$ is set as the Euclidean norm in RADTD. Consequently, the rescaled time series $D'$ can be converted into a sequence of URPs $R = \{R^1, R^2, ..., R^i, ..., R^T\}$. The URP subsequence length is $k(k < T)$; thus, RADTD can analyze the dynamical features of the time series through every URP's subsequence $R_i = \{R^i, R^{i+1}, ..., R^{i+k-1}\}$.

#### 2) ELM-AE TRAINING MODULE

After the URP image conversion process, RADTD applies the ELM-AE to learn the hidden trend or pattern within the URP sequence.

In ELM-AE, URPs in $i$-th subsequence $R_i = [R^i, R^{i+1}, ..., R^{i+k-1}]^T$ are mapped to an $L$-dimensional space with random orthogonal weights of hidden nodes. As shown in Fig. 2, the hidden layer can be calculated by (4):

$$\vec{h}(R) = g(\vec{a}R + \vec{b}), \quad \vec{a}^T\vec{a} = I, \vec{b}^T\vec{b} = 1, R \in R_i \quad (4)$$

where $\vec{h}(R) = [h_1(R),\ldots,h_L(R)]$ are the hidden node outputs, and $\vec{a} = [\vec{a}_1,\ldots,\vec{a}_L]^T$ and $\vec{b} = [b_1,\ldots,b_L]^T$ are orthogonal weights and biases, respectively, which are generated randomly. The output function is given by (5):

$$f_L(R) = \sum_{p=1}^{L} \beta_p h_p(R) = \vec{h}(R)\vec{\beta} \quad (5)$$

where $h_p(R) \in \vec{h}(R)$ represents the $p$-th hidden node output, and $\vec{\beta} = [\beta_1,\ldots,\beta_L]^T$ represents the output weight matrix. As $R_i = [R^i, R^{i+1},\ldots,R^{i+k-1}]^T$ represents the input and the output data, ELM-AE's output weight $\vec{\beta}$ is responsible for learning the transformation from the feature space to input data and satisfy (6):

$$\underset{\vec{\beta}}{Minimize} \left\| H\vec{\beta} - R_i \right\|^2 \quad (6)$$

where $H = [\vec{h}(R^i), \vec{h}(R^{i+1}),\ldots,\vec{h}(R^{i+k-1})]^T$. Thus, the output weights $\vec{\beta}$ can be calculated by (7):

$$\vec{\beta} = H^\dagger X \quad (7)$$

where $H^\dagger$ represents the generalized inverse matrix of the output matrix $H$. To enforce the result's robustness, a regularization term is added in the output weight's $\vec{\beta}$ calculation by (8):

$$\vec{\beta} = (\frac{I}{C} + H^T H)^{-1} H^T X \quad (8)$$

Hence, the output weights $\vec{\beta}$ learn to project from the hidden layer to the URPs themselves and can be applied in URP reconstruction.

### 3) SELF-SETS MODULE

The self-sets module is responsible for storing learned dynamical features in the form of ELM network parameters, including input weights $\vec{a}$, input biases $\vec{b}$, and output weights $\vec{\beta}$. The parameter set of each ELM training $\{\vec{a},\vec{b},\vec{\beta}\}$ is independently stored in self-sets as a pattern. As normal patterns are relatively consistent, multiple patterns in self-sets can be merged with the similarity threshold $\varepsilon$. Hence, patterns in self-sets and threshold represent the range of normal patterns, and patterns far away from self-sets can be viewed as abnormal.

### 4) ELM TESTING MODULE

Assuming the time series constantly updates and the latest untested subsequence is $D_N = \{\vec{x}_N, \vec{x}_{N+1},\ldots,\vec{x}_{N+w-1}\}$, RADTD reconstructs the URP of $D_N$ based on self-sets. $D_N$ is first converted into a URP $R^N$ following the image conversion module. Given the self-sets $\{P_1,\ldots,P_T\}$ and the $i$-th pattern $P_i = \{\vec{a},\vec{b},\vec{\beta}\}_i$, each parameter set in the self-sets can form an ELM network to reconstruct the untested URP $D_N$. Then, the ELM figures out the reconstructed URP $\hat{R}^{N,i}$ by (4) and (5), which represent the URP under pattern $P_i$. Corresponding to self-sets $\{P_1,\ldots,P_T\}$, the reconstructed URPs of URP $R^N$ are recorded as $\{\hat{R}^{N,1},\ldots,\hat{R}^{N,T}\}$.

### 5) ANOMALY RECOGNITION MODULE

Eventually, the original URP $R^N$ is compared with the reconstructed URPs $\{\hat{R}^{N,1},\ldots,\hat{R}^{N,T}\}$ to recognize the abnormal pattern. The difference between the two URPs is computed by reconstruction error, and RADTD is applied to the radial basis function in computing the reconstruction error $S_{N,i}$, as shown in (9):

$$S_{N,i} = 1 - exp(\frac{-\left\| R^N - \hat{R}^{N,i} \right\|}{2}) \quad (9)$$

where $\left\|\cdot\right\|$ is the Euclidean norm. As each reconstructed URP represents a pattern, a higher reconstruction error means the pattern of untested URP is different from the stored pattern. Hence, the anomaly score $S_N$ is equal to the minimum of reconstruction errors, which means the difference with the most similar pattern.

RADTD sets $\varepsilon$ as the threshold of anomaly score to identify anomalies. In a small amount of time series, the threshold needs to be set manually. As the data size increases, the distribution of the anomaly scores is closed to a normal distribution, and $\varepsilon$ can be determined based on its confidence interval.

If the minimal reconstruction error $S_N > \varepsilon$, the untested URP $R^N$ cannot match any pattern in self-sets, and the updated time series $D_N$ is referred to as abnormal. Otherwise, the untested URP's pattern is regarded as similar to one stored pattern in self-sets.

## IV. BENCHMARK DATASETS VERIFICATION

This section describes how the researchers conducted experiments of proposed framework RADTD with the Yahoo Webscope anomaly detection benchmark. The results were compared with the ELM-AE and three conventional RQA indicators to check the method's performance.

### A. DATASETS

The Yahoo Webscope anomaly detection benchmark supports real and synthetic univariate time-series problems for anomaly detection [40]. It consists of a real dataset (A1 Benchmark) and three synthetic datasets (A2 Benchmark, A3 Benchmark, and A4 Benchmark) with 367 instances in total, shown in Table 1. The A1Benchmark dataset describes the production traffic in Yahoo traffic services with 67 datasets. The number of points for each dataset in the A1Benchmark ranges from 741 to 1461 observations recorded at regular intervals, and anomalies are manually labeled. Other three datasets separately include 100 datasets with anomalies that are generated algorithmically.

### B. EXPERIMENT DESIGN

In this section, the proposed method (RADTD) was verified by the Yahoo benchmark datasets for anomaly detection. For comparison, two involved techniques, ELM-AE and RP, were also separately applied for anomaly detection. The ELM-AE method in the experiment learns and reconstructs the embedding of time-series subsequences by ELM-AE, and detects the abnormal ones with a high reconstruction error. The common method for RP is RQA, which can quantify the RP's structure with complex measures. In this experiment, three common RQA indicators were chosen for anomaly detection: recurrence rate (RR), determinism (DET), and laminarity (LAM). RQA indicators with high difference were regarded as anomaly occurrences.

Experiments contained two aspects for verification, as shown in Table 2. In accuracy experiments, RADTD, the ELM-AE method, and RQA indicators detected all the 367 time series in Yahoo benchmark datasets. In the robustness experiments, randomly selected datasets were examined by RADTD, along with RQA indicators under different RP thresholds.

The performance of methods was evaluated by area under curve (AUC), which is commonly used for unbalanced datasets. The higher the AUC score, the better the performance of the method.

Related parameter setting in this experiment is listed in Table 3. RADTD and ELM-AE chose the sigmoid function $g(x) = 1/(1+e^{-x})$ as the activation function in the input project function. Considering the influence of the RP threshold $\epsilon$ on RQA measure, the experiment calculated the AUC for the RQA measures with the varying threshold of $\epsilon \in [0,5]$, and observed the optimal $\epsilon$ with the maximum AUC.

### C. EXPERIMENT RESULTS AND ANALYSIS

Experimental results of the three methods in terms of accuracy and robustness were analyzed for 367 Yahoo benchmark datasets.

#### 1) RESULTS OF ACCURACY EXPERIMENT

Fig. 6(a) presents the average AUC score and ranking among different methods in the whole and each individual benchmark. Each histogram is in descending order by the AUC score. The total average AUC scores of all methods are above 0.900, and RADTD is slightly ahead of the other methods. The AUC scores of different methods are different among the benchmark datasets. ELM-AE has the highest AUC score in the A1Benchmark, but it has bottom rankings in the other benchmark datasets. RQA-DET and RQA-LAM also possess such ranking contrasts among the benchmark datasets. However, the AUC score of RADTD ranks in the top two for any benchmark dataset.

Fig. 6(b) counts the proportion of each method ranking first in 376 instances. RADTD ranks first in 35% of the instances, and RQA-RR leads in 27% of the instances. Other methods are ahead in less than 20% of the datasets.

To compare these methods' performance in each instance, Fig. 6(c) shows the AUC score of five methods in 367 instances by four radar maps, which are in descending order by the AUC score of RADTD. As shown in Fig. 6(c), RADTD occupies the periphery of the radar map with a higher AUC score. With the decrease of RADTD's AUC score, the scores of other methods constantly fluctuate and mainly exist inside the RADTD circle.

Fig. 7 depicts the dispersion of the benchmark dataset by boxplots. Among the four box plots, RADTD maintains a higher median and a narrower inter-quartile range in each benchmark, outperforming other methods. The AUC score of RQA-LAM is slightly weaker than RADTD overall. RQA-DET shows unstable results overall, with a larger inter-quartile range. ELM-AE outperforms RADTD in the A1Benchmark, while in other benchmarks, it may not be competitive enough. Moreover, differences among benchmarks are relatively obvious: the AUC scores of A2benchmark for all methods are high and very close; and A4benchmark is more difficult to detect; the average AUC of each method is generally lower than that of other benchmarks, and every inter-quartile range is wide.

In summary, Fig. 6 and 7 reflect the difference in accuracy between RADTD and RQA and ELM-AE for different data. The accuracy of RADTD was generally better than that of the ELM-AE method, which indicates that URP can highlight the dynamics of time series and thus help to detect time series anomalies. Compared with RQA indicators, RADTD performed better for different data and was more stable for different data. There are three reasons for the differences in accuracy between the three methods: First, the RP conversion successfully represents dynamical features, meaning ELM-AE performed poorly compared with others in most datasets. Second, the URP with the removed threshold avoids information loss, resulting in higher accuracy in RADTD than those in RQA indicators. Besides, the ELM-AE in RADTD learned enough dynamical features, further highlighting the nuance among time-series. Therefore, RADTD improves the accuracy of dynamical feature-based anomaly diagnosis by combining both ELM-AE and URP methods.

#### 2) RESULTS OF ROBUSTNESS EXPERIMENT

In the accuracy experiments, the accuracy of RADTD and RQA indicators were very close. However, the RP thresholds of the RQA indicators were optimized, and the results of RQA were influenced by the RP thresholds. Therefore, robustness experiments were conducted to analyze the RP thresholds of RADTD and the RQA indicators.

Fig. 8 shows the AUC scores of RQA indicators in four instances under different RP thresholds, and the RADTD results are also included for comparison. The maximum AUCs for each instance is shown in Table 4. For A1Benchmark-TS15, the maximum AUCs of RQA-LAM, RQA-RR, and RQA-DET are 0.997, 0.990, and 0.950 at $\epsilon = 1.4$, $\epsilon = 1.4$, and $\epsilon = 1.3$, respectively (dotted line). RADTD omits the limitation of RP threshold, and its AUC score remains at 0.990. An AUC level of 0.5 (dashed-dotted line) indicates that the RQA measure is unable to detect anomaly points. The AUCs of RQA indicators vary obviously with the RP threshold increase from 0.1 to 5. Generally, the maximum AUCs of RQA measures are close or even higher than the AUC of RADTD. Except for A2Benchmark-TS18, however, RQA indicators attain the maximum AUCs in rare cases, and drop into lower AUC levels at most RP thresholds.

In summary, as seen in Fig. 8 and Table 4, the results of RQA were strongly limited by the actual RP thresholds. Under the optimal RP threshold, the maximum AUC value of RQA was very close to that of RADTD, and as shown in Fig. 8,

RQA achieved a high AUC only in a very narrow range of thresholds. The AUC was at a low level in other wider threshold ranges. In contrast, RADTD was not affected by RP thresholds and detects anomalies as consistently, and it avoided the time-consuming threshold optimization. Therefore, RADTD possesses better robustness than RQA indicators.

## V. TUNNEL ENGINEERING SIMULATION EXPERIMENTS
### A. ENGINEERING BACKGROUND
Abnormal soil settlement events may indicate serious engineering risks during the construction of tunnel engineering project. The construction principle of tunnel engineering is shown in Fig. 9. Soil settlement can be accurately measured by instruments at the surface; however, surface instruments have limited numbers with low frequency. Besides, there is always a delay in reflecting substantial changes in the soil to the surface, making it difficult to detect abnormal settlement changes in time. Current research shows that the excavation and the slurry chamber pressures are highly correlated with soil pressure and are more sensitive to the settlement change [41, 42]. Therefore, with pressure sensors on shield machines, detecting abnormal pressure change is a common way to warn against the engineering risk of a shield machine. In this study, pressure parameters were used to detect abnormal soil settlement events.

### B. DATASETS AND EXPERIMENT DESIGN
Table 5 shows the three datasets that were collected from real-world tunnel engineering applications. All of the datasets were provided to detect abnormal soil settlement events during the construction process.

The River Tunnel dataset is about the C Shanghai Tunnel project. According to the major event record from tunneling construction, the ground surface suffered quite a severe settlement during construction at 14:00 on 2008-05-19. The dataset collected excavation chamber pressure data from a sensor related to the accident per 3-minute interval.

The Subway Tunnel dataset is about a sensor in the Y Shanghai subway tunnel project, which monitored the excavation chamber pressure data in 4 months at a 3-minute interval. During the excavation of the tunnel, the critical tunnel settlement occurred on 2018-05-17 03:06.

The Road Tunnel dataset contains pressure data from Wuhan S Road Tunnel project. Five slurry chamber pressure sensors were applied to monitor the change of pressure, and a critical tunnel settlement risk occurred on 2017-11-01.

Based on the above three different projects datasets, this study conducted three simulation experiments to detect abnormal events in three projects. By comparing with the anomaly time recorded by the project, this experiment analyzed the accuracy of RADTD in detecting anomalies. The robustness of the model in different datasets was also verified by setting the same parameters for each project in the simulation experiments. The parameter settings are shown in Table 6, where the abnormality threshold is set according to the expert opinion as $\varepsilon = 0.99$.

### C. SIMULATION IN RIVER TUNNEL PROJECT
Fig. 10 presents the excavation chamber pressure curve of the River Tunnel dataset. It is obvious that the excavation chamber pressure of the River Tunnel dataset goes smoothly at first and then takes a downward trend. The major event record of the tunneling construction site stated that during construction, the ground surface experienced a severe settlement at 14:00 on 2008-5-19, which is marked by a red star in Fig. 10. RADTD reported anomalies from 2008-5-15 10:54 to 11:00 (red line), which means the approach detected the anomaly 4 days before the ground settlement occurred.

In Fig. 10, RADTD alarms 4 days earlier than the actual recorded anomaly. Moreover, by the post-analysis of the responsible construction personnel, the soil layer of the shield project was already in an unstable state, with the risk of collapse at 10:54 on 2008-5-15, and this situation continued to deteriorate afterwards, culminating in the collapse accident on 2018-5-19. Therefore, the simulation detection results of the RADTD model are accurate and can report anomalies for a longer period of time in advance to achieve the effect of early warning.

### D. SIMULATION IN SUBWAY TUNNEL PROJECT
The result of dataset Subway Tunnel dataset is shown in Fig. 11. Under normal working conditions, the pressure of the earth bunker is relatively stable, fluctuates in the range of 4.5–6 during propulsion, and is also stable at about 6 during assembly. RADTD reports the anomaly for 4 times: 2018-05-17 1:18, 3:09, 3:12, and 9:75, where red line highlights; and the soil settlement was measured and recorded at 2018-05-17 09:00 in the morning, which is marked by star.

Referred to the engineering regulation, the Subway Tunnel project conducted a routine settlement check at 9:00 and 17:00 every day and reported a single large anomalous settlement during the surface inspection on 2018-5-17 at 9:00. The model detected the abnormal change in earth pressure as early as 1:00 on May 17 and reported the abnormality three times before the routine check, warning the settlement risk of the model. The fluctuations of soil pressure at 1:00-3:00 on 2018-5-17 were analyzed by technicians afterward, and there was a risk of collapse. This shows that RADTD model monitoring can monitor settlement and detect abnormal changes in construction earlier than routine inspections.

### E. SIMULATION IN ROAD TUNNEL PROJECT
The result of Road Tunnel dataset is shown in Fig. 12. The marked area shows the detected abnormal time series subsequence, and the star highlights the recorded anomaly timestamp. In the Road Tunnel dataset, the mud and water pressure monitored by five sensors remained largely stable during daily working conditions. The incident time marked in the construction records also occurred around 2017-11-1 4:27, as indicated by the asterisk. The RADTD model run found soil pressure anomalies at five sensors during 2017-11-1 4:27-4:42 and 2017-11-2 1:03-1:12, as indicated by the red line.

According to the detailed description of the engineering anomaly record, the event in the Road Tunnel was an unexpected event, and the incident occurred when the construction crew stopped the advancement and found that some sub-equipment failed, which triggered the unstable working status of the shield. Further, the RADTD model detected first and reported the risk event of ground instability under the influence of equipment failure.

*F. ANALYSIS FOR TUNNEL ENGINEERING SIMULATIONS*

Since three real-world engineering projects with anomalous records were selected in simulation experiments, the accuracy was measured by the recorded anomaly time and professional analysis. As can be seen from the RADTD alarm time and recorded anomaly time in Fig. 10–12, the first alarm time is at or earlier than the anomaly recorded time in each dataset. In the River Tunnel dataset and the Subway Tunnel dataset, RADTD can detect the abnormal change of soil pressure early than the recorded ground settlement anomaly time point. Even in the Road Tunnel dataset which contains an emergency, RADTD can detect the anomaly at the first time. Three datasets indicate that RADTD has high accuracy in practical application.

In addition, this study mainly analyzed the robustness through the parameter-setting among different projects. RADTD detected anomalies of pressure time series in three projects with the same parameters. The detection results indicate that this experiment arrangement has no effect the early warning of all abnormal events by RADTD, which means the high robustness of RADTD.

## VI. CONCLUSION

The real-world application requires a time-series anomaly detection method that detects anomalies in time with less priori knowledge and human intervention. In this study, we put forward a framework, RADTD, to detect time-series anomalies automatically with less label dependency and high robustness. RADTD represents dynamical features with the URP method, learns automatically by the ELM-AE network, and recognizes anomalies with the assistance of negative selection.

This article contains three contributions. First, RADTD effectively combines negative selection, URP, and ELM-AE, which can automatically represent and learn time-series features in the perspective of dynamics, and recognize anomalies under changeable patterns with updatable self-sets. Second, compared with respectively ELM-AE method and RP method, RADTD is more accurate and robust in detecting anomalies in complex time series. Finally, the simulation experiments on tunnel engineering datasets show its potential value for the early warning of engineering risk and further application to other situations which have difficulty in labelling and are urgent for accurate detection.

A follow-up of this framework will focus on three points: First, stored patterns in self-sets are clustered by a similarity threshold, which can be optimized further. Second, parameters such as window length and step are vital to URP representation, and related parameter setting remains to be researched in industry applications. In addition, with the aid of engineering expertise, we will focus on the correspondence relationship of the detailed anomaly pattern and the specific fault type to establish a risk database for better understanding engineering disasters.

## ACKNOWLEDGMENT

This work was supported by Shanghai Science and Technology Commission's Plan: Study on optimization of intelligent identification and control strategy of shield driving status (No.18DZ1205502).

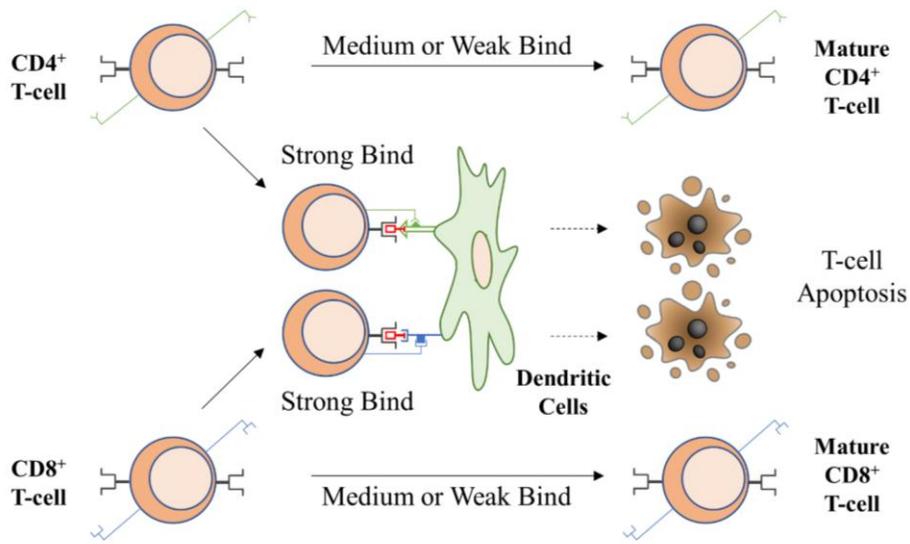

**Figure 1.** Negative selection process of T cell in biological immune immunity.

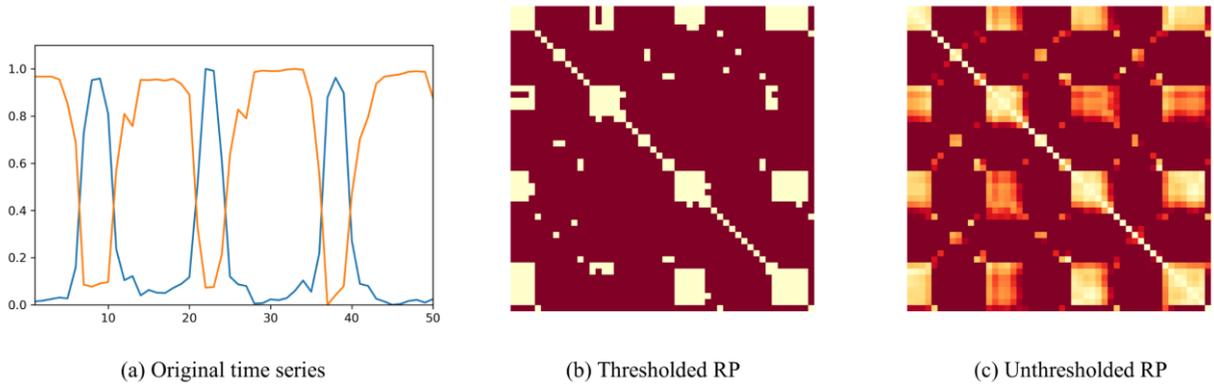

**Figure 2.** Multivariate time series (a), corresponding thresholded RP (b) and unthresholded RP (c).

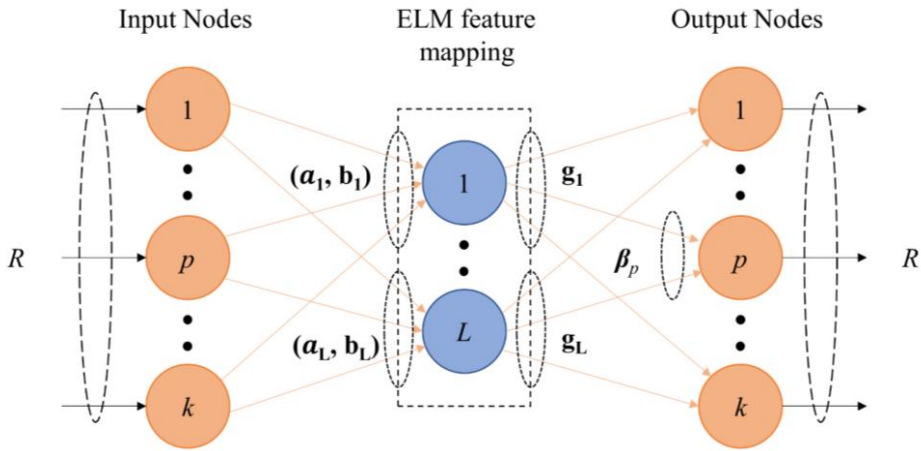

**Figure 3.** The diagrammatic sketch of ELM-AE.

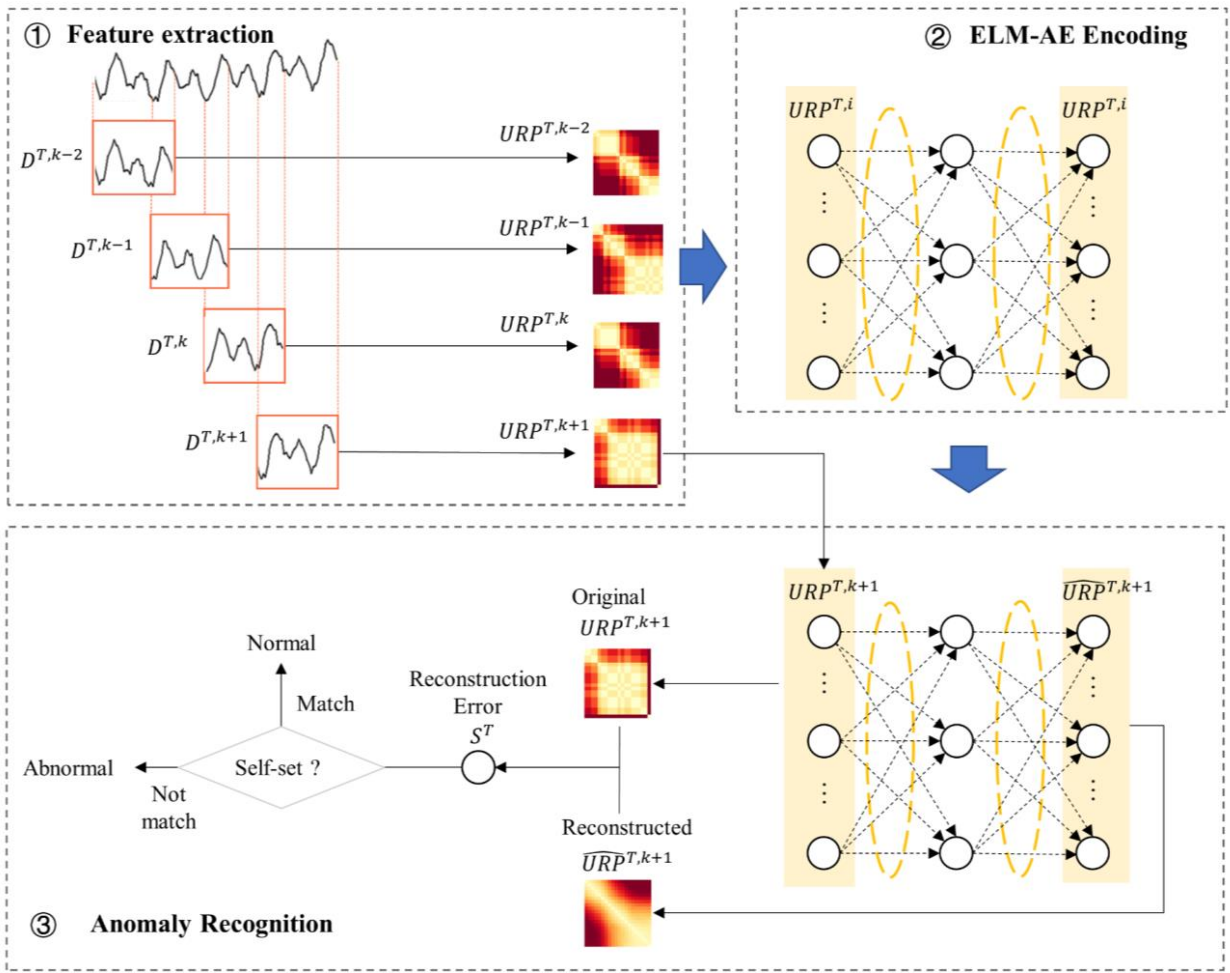

**Figure 4.** Three functions of time series data processing.

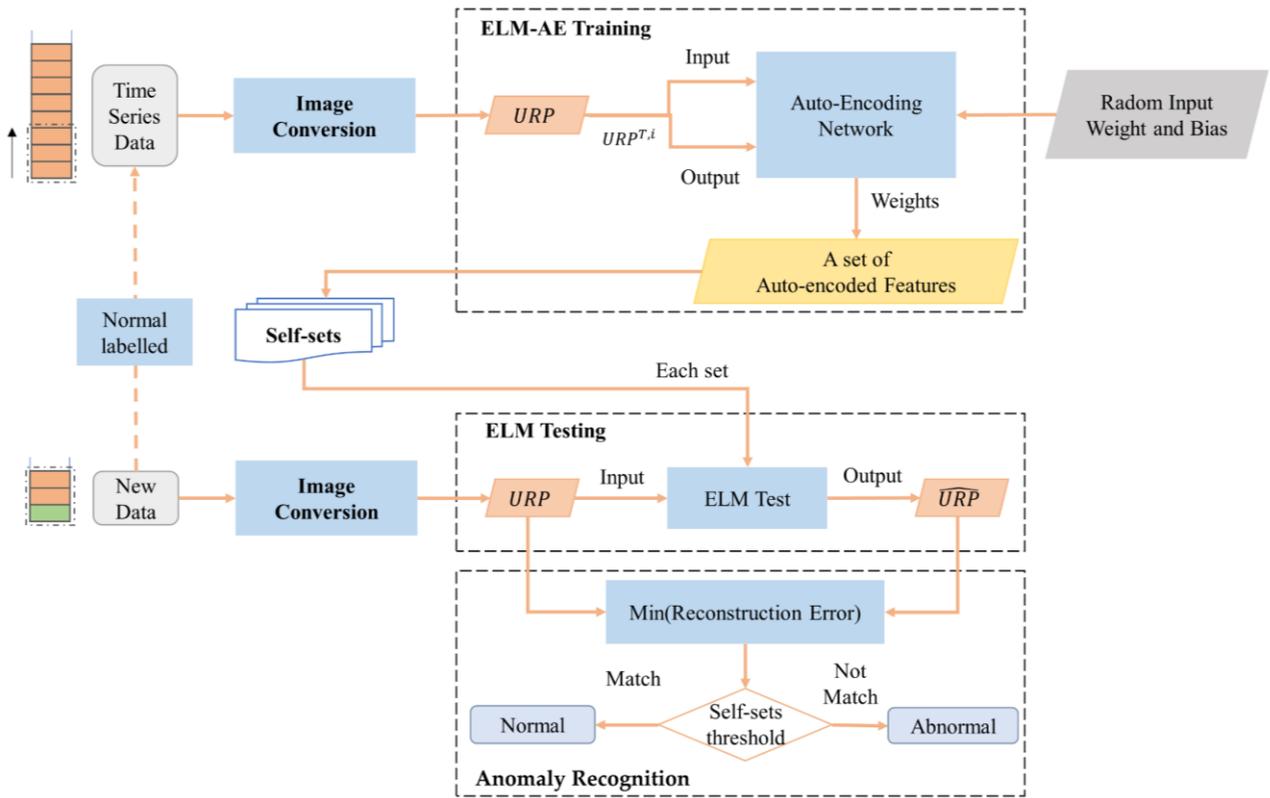

**Figure 5.** The flow chart of RADTD.

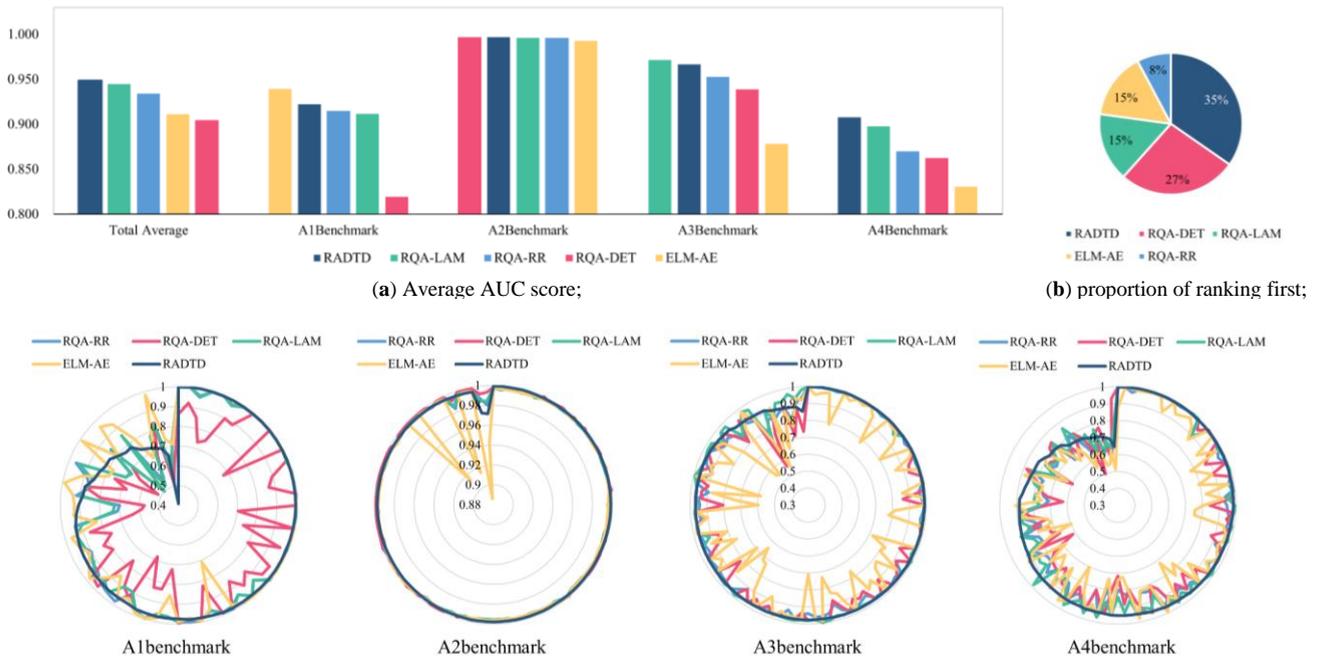

(**a**) Average AUC score; (**b**) proportion of ranking first;

(**c**) AUC radar map in each benchmark.

**Figure 6.** Average AUC of RADTD, RQA indicators, and ELM-AE on Yahoo datasets. (**a**) Average AUC score; (**b**) proportion of ranking first; (**c**) AUC radar map in each benchmark.

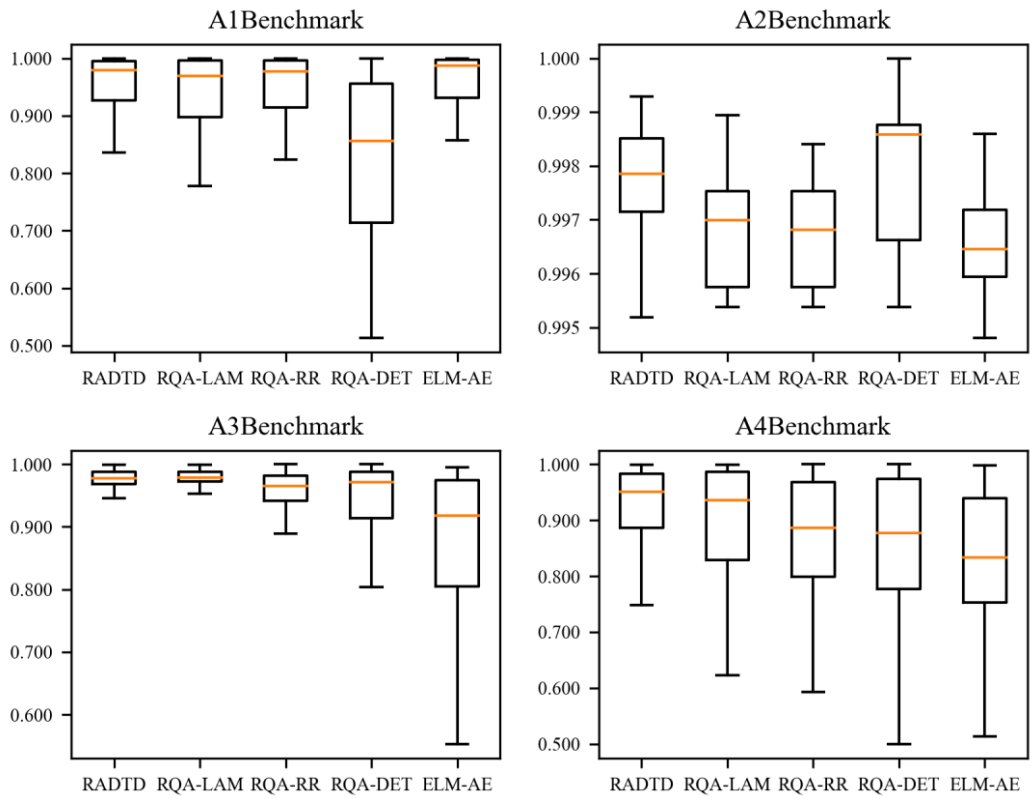

**Figure 7.** AUCs of RADTD, RQA indicators and ELM-AE within Yahoo benchmark datasets.

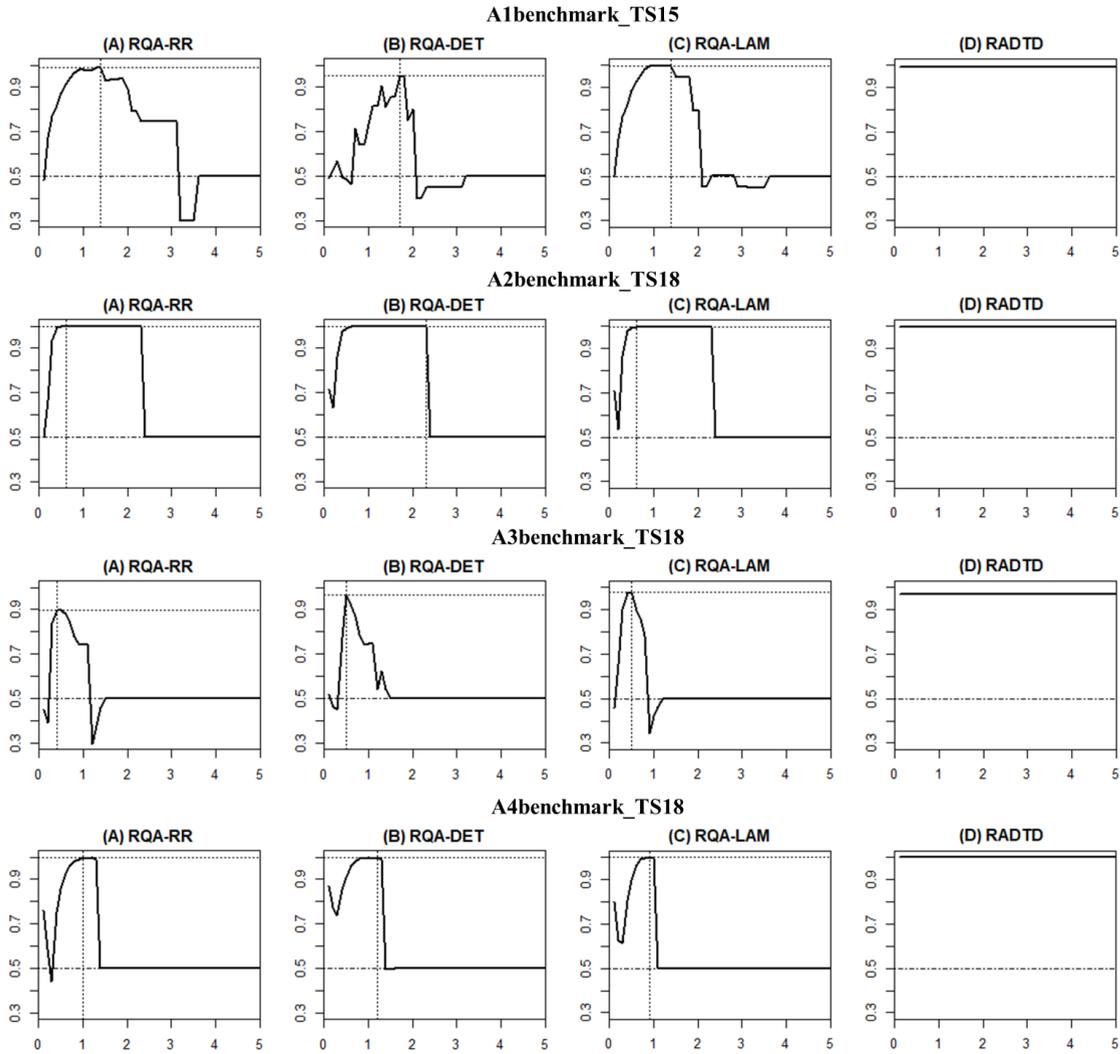

**Figure 8.** AUCs vs. RP threshold for the measures (A) RQA-RR, (B) RQA-DET, (C) RQA-LAM, and (D) RADTD for A1Benchmark-TS15, A2Benchmark-TS18, A3Benchmark-TS18, and A4Benchmark-TS18 instances.

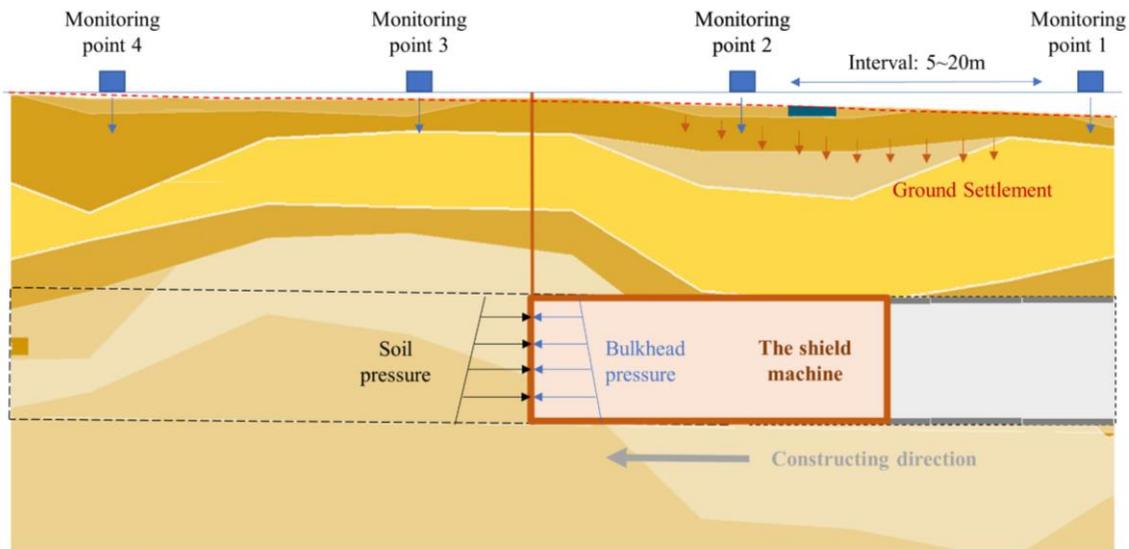

**Figure 9.** Schematic diagram of shield machine construction process.

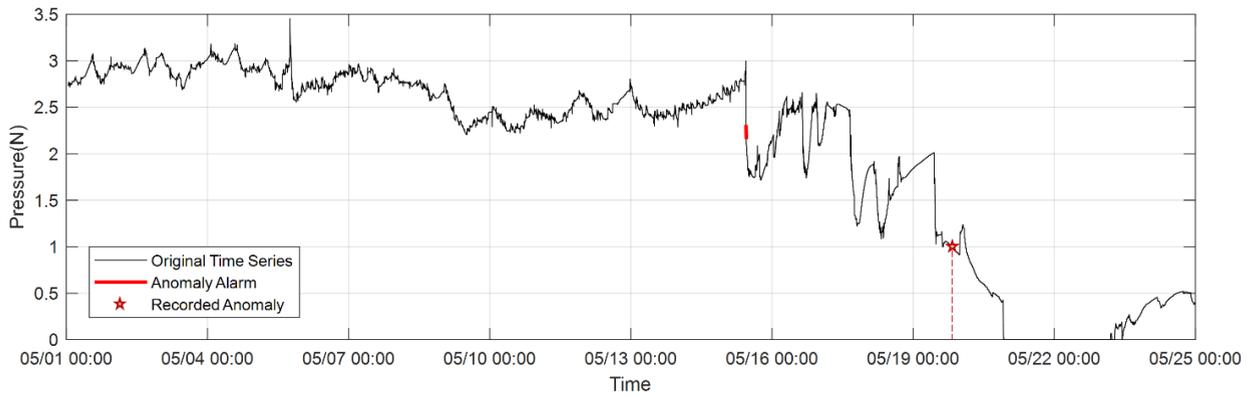

**Figure 10.** Results for River Tunnel dataset.

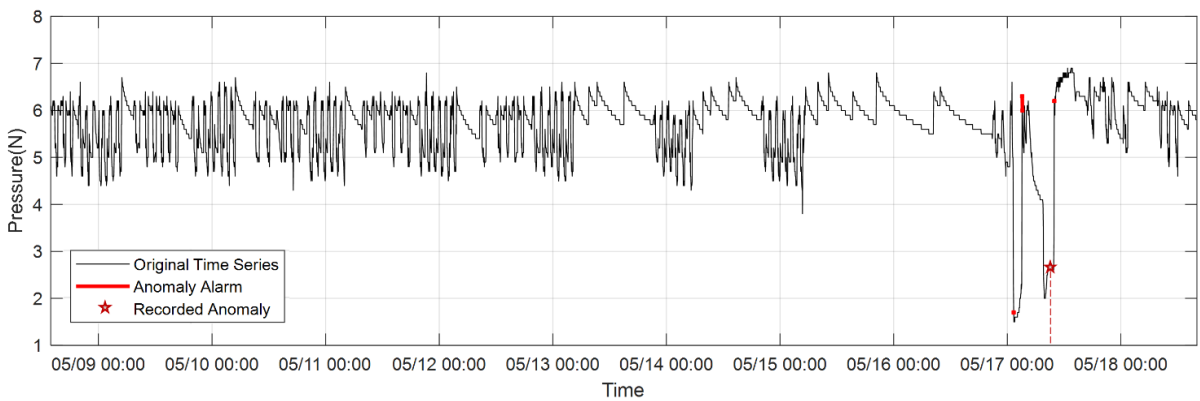

**Figure 11.** Results for Subway Tunnel dataset.

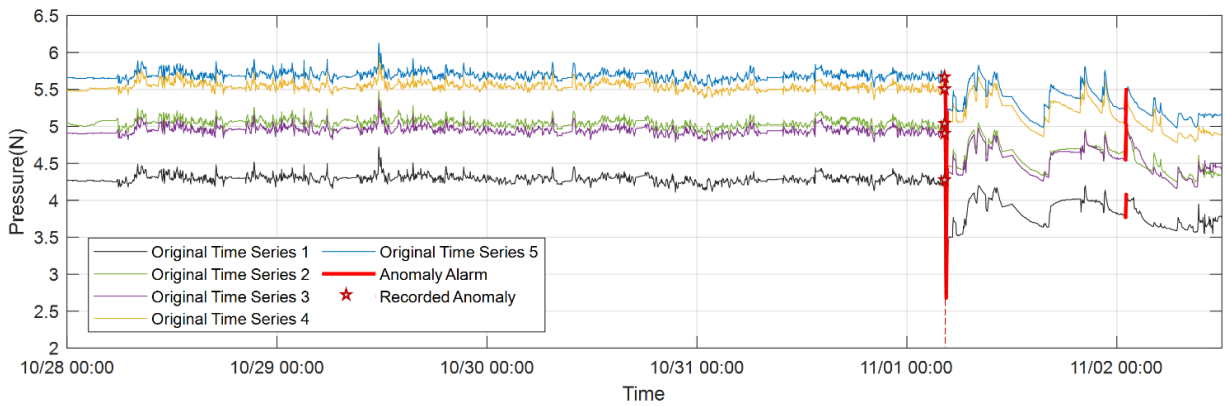

**Figure 12.** Results for Road Tunnel dataset.

**Table 1.** The description of datasets in experiments.

| Data Source | Datasets | Dataset | Size | Type |
|---|---|---|---|---|
| Yahoo Webscope (367 datasets) | A1 Benchmark | 67 (TS1-TS67) | 741-1461 | Univariate |
| | A2 Benchmark | 100 (TS1-TS100) | 1421 | Univariate |
| | A3 Benchmark | 100 (TS1-TS100) | 1680 | Univariate |
| | A4 Benchmark | 100 (TS1-TS100) | 1680 | Univariate |

**Table 2.** The experiment

| Experiment | Datasets | Involved methods |
|---|---|---|
| Part 1: Accuracy | All (367 datasets) | RADTD, ELM-AE, RQA |
| Part 2: Robustness | A1 Benchmark-TS15 A2 Benchmark-TS18 A3 Benchmark-TS18 A4 Benchmark-TS18 | RADTD, RQA |

**Table 3.** Main parameters used in benchmark experiments.

| Method | Parameter | Value |
|---|---|---|
| All methods | Length of the subsequence: $w$ | $w=15$ |
| All methods | Embedding dimension: $m$ | $m=1$ |
| RADTD | Length of the URP window: $t$ | $t=5$ |
| RADTD | Number of the input URPs: $k$ | $k=10$ |
| RADTD | Number of hidden nodes: $L$ | $L=10$ |
| RQA | Length of RP window: $t$ | $t=5$ |
| RQA | RP threshold in RP representation | Ranging from 0 to 5 and step by 0.1 |
| ELM-AE | Number of input nodes: $k$ | $k=10$ |
| ELM-AE | Number of hidden nodes: $L$ | $L=10$ |

**Table 4.** The maximum AUCs and the optimal $\epsilon$ in four instances.

| Datasets | RQA-LAM | | RQA-RR | | RQA-DET | | RADTD | |
|---|---|---|---|---|---|---|---|---|
| | AUC | $\epsilon$ | AUC | $\epsilon$ | AUC | $\epsilon$ | AUC | $\epsilon$ |
| A1 Benchmark-TS15 | 0.997 | 1.4 | 0.990 | 1.4 | 0.950 | 1.7 | 0.990 | - |
| A2 Benchmark-TS18 | 0.998 | 0.6 | 0.998 | 0.6 | 0.999 | 2.3 | 0.997 | - |
| A3 Benchmark-TS18 | 0.977 | 0.5 | 0.898 | 0.4 | 0.964 | 0.5 | 0.967 | - |
| A4 Benchmark-TS18 | 0.998 | 0.9 | 0.996 | 1 | 0.997 | 1.2 | 0.999 | - |

**Table 5.** The Description of Engineering Datasets.

| Project name | Datasets | Size | Type | Time period | Interval | Physical meaning |
|---|---|---|---|---|---|---|
| Shanghai C River Tunnel, China* | River Tunnel | 11721 | Univariate | 2008-05-01 to 2008-05-25 | 3 min | Excavation chamber pressure |
| Shanghai Y Subway Tunnel, China* | Subway Tunnel | 49066 | Univariate | 2018-01-22 to 2018-05-18 | 3 min | Excavation chamber pressure |
| Wuhan S Road Tunnel, China* | Road Tunnel | 149521 | Multivariate | 2017-03-09 to 2017-11-02 | 3 min | Slurry chamber pressure 1-5 |

* The datasets come from three tunnel projects in China, but restrictions apply to the availability of the data, which were used under license for the current study but are not publicly available.

**Table 6.** Main parameters used in simulation experiments.

| Method | Parameter | Value |
|---|---|---|
| RADTD | The length of subsequence: $w$ | $w=15$ |
| RADTD | The embedding dimension: $m$ | $m=1$ |
| RADTD | The length of URP window: $t$ | $t=5$ |
| RADTD | The number of input URPs: $k$ | $k=10$ |